# Gender Differences in Abuse: The Case of Dutch Politicians on Twitter

Isabelle van der Vegt
i.w.j.vandervegt@uu.nl

Department of Sociology, Utrecht University

**ABSTRACT**

Online abuse and threats towards politicians have become a significant concern in the Netherlands, like in many other countries across the world. This paper analyses gender differences in abuse received by Dutch politicians on Twitter, while taking into account the possible additional impact of ethnic minority status. All tweets directed at party leaders throughout the entire year of 2022 were collected. The effect of gender and ethnic minority status were estimated for six different linguistic measures of abuse, namely, toxicity, severe toxicity, identity attacks, profanity, insults, and threats. Contrary to expectations, male politicians received higher levels of all forms of abuse, with the exception of threats, for which no significant gender difference was found. Significant interaction effects between gender and ethnic minority status were found for a number of abuse measures. In the case of severe toxicity, identity attacks, and profanity, female ethnic minority politicians were more severely impacted than their ethnic majority female colleagues, but not worse than male politicians. Finally, female ethnic minority politicians received the highest levels of threats compared to all groups. Given that online abuse and threats are reported to have a negative effect on political participation and retention, these results are particularly worrying.

## Introduction

In 2022, 1,125 reports of threats made to politicians were filed to the Dutch police (Openbaar Ministerie, 2023). This figure stands in stark contrast with the 588 reports filed in 2021, and 200 in 2015 (Jonker, 2022). A survey with members of the House of Representatives in the Netherlands showed that 12% has kept an opinion to themselves due to fear of threats or intimidation (Jonker & van der Parre, 2022). While a large number of threats are made online, some threateners also physically approach politicians. In January 2022, the Dutch minister of Finance was approached at her home by a man shouting conspiratorial slogans while carrying a burning torch (NOS Nieuws, 2022). For women in politics, the issue appears to be particularly salient. Such threats not only put individual women at risk, but they may also have a negative effect on women's participation in politics as a whole. In 2019, eighteen female UK MPs stood down from their seat, with several of them announcing that the abuse they received was a factor in their decision (Scott, 2019). A study drawing on interviews with 101 Canadian politicians showed that online harassment did not necessarily stifle political ambition, but the hostile work environment reportedly affected the ability of these politicians to do their job and their willingness to stay in the job (Wagner, 2022). Worries of family members as a result of threats are reportedly one of the main reasons for people to leave politics in the Netherlands (NOS nieuws, 2023). Considering the growing trend of online abuse directed at female politicians and its far-reaching consequences, the current study empirically analyses gender differences in online abuse received by Dutch politicians on Twitter, while taking into account the growing body of evidence that ethnic minority women are particularly affected. This contribution offers several different measures of abuse and a broad timeline of tweets directed at Dutch politicians throughout the entire year of 2022.

## Background

The following sections cover empirical evidence on the prevalence of online abuse and gender differences herein, followed by theoretical explanations of online abuse, gendered abuse of politicians, and the abuse of minority ethnic politicians.

### *Prevalence of (gendered) online abuse*

Several survey studies have examined the levels of online harassment, abuse and threats experienced by politicians. Surveys of MPs from the early 2000s already showed that 10% of UK MPs experienced unwanted contact via social media in 2010 (James et al., 2016), to 60% of New Zealand MPs in 2014 (Every-Palmer et al., 2015). More recent surveys report a further rise in abusive online messages in recent years, with increases from 40% in 2013 to 70% in 2021 in Norway (Bjørgo et al., 2022). Others have shown that 100% of MPs in the UK (Akhtar & Morrison, 2019) and Victorian (Phillips et al., 2023) parliament reported experiencing abuse via social media, in 2018 and 2021, respectively. Similar patterns have been reported when examining social media data. Gorrell et al. (2020) studied Twitter data surrounding the 2019 parliamentary elections in the UK and observed that 4.46% of replies to MPs could be considered abusive, compared to 3.27% in the same period in 2017.

Case studies leveraging social media data have clearly demonstrated the severity of online abuse and threats experienced by female politicians. Examples include a study of abusive language and hate speech directed at Japanese female politicians (Fuchs & Schäfer, 2021), and sexual, physical, and psychological threats on Twitter directed at pro-choice female politicians in Chile (Pérez-Arredondo & Graells-Garrido, 2021). However, empirical evidence on gender differences in online abuse and threats received by politicians is mixed. The majority of studies that have made direct comparisons between male and female politicians focus on the UK. Ward and McLoughlin (2020) report that male UK MPs received significantly more abusive tweets than female MPs in the UK between November 2016 and January 2017. However, female MPs received more hate speech than their male counterparts, partially explained by the inclusion of gendered slurs as hate speech. Gorrell et al. (2020) similarly found that male politicians received more general and political abuse, while women received more sexist abuse. Southern and Harmer (2021) collected 117,802 tweets directed at UK MPs from two weeks in 2018 to assess gender differences. They found that female MPs received significantly more incivility, and were more likely to receive tweets that stereotyped them or questioned their position as representatives. Esposito and Breeze (2022) examined tweets for three weeks preceding the UK general elections in December 2019, and found no marked semantic differences (i.e., linguistic measures of emotion, references to appearance and intelligence) in tweets directed at male and female MPs. However, they did find large inter-individual differences, in that certain female MPs received a disproportionate amount of tweets containing words referring to appearance, sexual relations, emotions, and violence (Esposito & Breeze, 2022).

Two noteworthy studies have empirically examined this question within the Dutch context. Tromble and Koole (2020) studied tweets directed at Dutch politicians in October 2013, and found a marginal effect for gender in predicting tone (i.e., negative/positive language in tweets). Dutch, female, non-populist politicians received somewhat 'friendlier' tweets than their male counterparts. The authors also qualitatively examined negative tweets, and found very few tweets that could be considered sexist (e.g., mocking a female politician's appearance or using gendered slurs) and even fewer that could be considered a physical threat. A second investigation published in weekly opinion magazine *De Groene Amsterdammer* examined tweets directed at female Dutch politicians from October 2020 to February 2021 (Saris & van de Ven, 2021; Veerbeek, 2021). Tweets were manually labelled as being hateful, threatening, or non-problematic. The hateful and threatening tweets were secondarily labelled for containing comments about (negative) female stereotypes, age, appearance, ethnicity, and religion. Thereafter, machine learning models were trained on this data and used to classify unseen tweets. The authors found that 10% of tweets directed at female politicians could be classified as hateful or threatening. Furthermore, it was found that female politicians were more frequently addressed with their first name and terms such as 'lady' and 'girl' ('vrouwtje' and 'meid' in Dutch) than their male counterparts with male equivalent terms.

*Explaining online abuse*

The online disinhibition effect, as described by Suler (2004), refers to the tendency of individuals to behave in ways online that they would not in person, due to the anonymity and minimization of authority on social media platforms, among other factors. This can lead to a sense of detachment from social norms, which may result in "acting out" online, for example in the form of harassment and threats directed at politicians (Trifiro et al., 2021; Tromble & Koole, 2020). Gorrell et al. (2020) proposed a heuristic framework specifically aimed at understanding the abuse politicians receive on Twitter. The framework consists of four factors which may explain the distribution of abuse among politicians, specifically, 1) prominence: abuse focuses on individuals most in the public eye, 2) event surge: specific political or media events may lead to a surge in attention and/or hostility towards a politician, 3) engagement: abuse may be the result of a specific tweet by the politician themselves, and 4) identity: personal characteristics of a politician may affect the abuse they receive. Gorrell et al. (2020) found support for their model of abuse in their dataset of tweets from 2019, with the majority of abuse focusing on high profile politicians. Specific events and engagement, such as television appearances and opinionated tweets, were associated with 'spikes' in levels of abusive. Finally, Gorrell et al. (2020) found that certain personal characteristics, such as political stance and gender, led to differences in the level and forms of abuse received.

*Explaining gendered online abuse*

Differences in prevalence and/or nature of abuse directed at female versus male politicians can be understood by leveraging theoretical frameworks from political science aimed at explaining violence directed at women in politics. Krook and Sanín (2020) suggest that violence against women in politics stems from misogyny, defined as "a system that polices and enforces patriarchal norms and expectations" (p. 742). The authors propose a three-factor model explaining violence against women in politics. First, the origins of this violence are structural, resulting from longstanding political theories associating men with the public and women with the private sphere. This, in turn, inspires and rationalizes hostility against female leaders, because they violate aforementioned female gender roles by operating in politics, a traditionally male-dominated domain. Second, Krook and Sanín (2020) put forward cultural violence as the means by which violence against women in politics is perpetrated. Specifically, cultural violence is the phenomenon of tolerating violence when it is directed at a particular group. Examples of this in the context of violence towards women in politics include sexist jokes and sexual objectification. Third, symbolic violence is seen as the (intended) outcome of violence against women in politics. This takes the form of male domination, aimed at "putting women who deviate from prescribed norms back in in their place" (p. 743). Bardall et al. (2020) put forward a related theoretical framework for gendered political violence. They separate gendered motives, forms, and impacts in order to distinguish between political violence and *gendered* political violence. Gendered motives are present when political violence is committed in order to preserve politics as a male domain. Gendered forms constitute the way in which violence is perpetrated using gendered roles, such as sexualized language. Finally, gendered impacts refer to the possible gender differences in the meaning and consequences of political violence. Bardall et al. (2020) mention the way in which the media and society narrate an incident of political violence and possible differences in the extent to which political violence causes women versus men to retreat from politics, as examples of gendered impacts.

*Online abuse of (ethnic) minorities*

In their theoretical framework of gendered political violence, Bardall et al. (2020) further specified that this form of violence can serve to preserve political power of the hegemonic male group. That is, politicians not belonging to the dominant cultural, ethnic, or religious group (e.g., gay men or ethnic minority women) may be similarly or even more strongly impacted than female politicians with majority characteristics. Indeed, various investigations into the gender differences associated with online abuse directed at politicians have found that ethnic minority women were targeted disproportionately. As a consequence, Esposito and Breeze (2022) proposed that 'gender bias is activated by some women more than others' (p. 320). Analysis conducted by Amnesty International on tweets directed at female MPs in the UK in 2017 showed that 41% of all abusive tweets directed at women MPs were sent to Black, Asian and Ethnic minority (BAME) women MPs, even though almost 89% of the sample of MPs were white (Amnesty International UK, 2017). A disproportionate amount (32%) of abusive tweets were directed at Diane Abbott, the first black woman MP. The authors note that abuse sent to her often included threats of sexual violence and frequently mentioned her gender and race. In the Dutch investigation of Saris & van de Ven (2021) it was also found that female politicians of color or those with minority religious beliefs were impacted more strongly. For example, up to 30% of tweets directed at a Muslim second chamber member were classified as abusive, versus the 10% average for female politicians in the sample.

## The current study

In explaining online abuse towards female politicians, both Krook and Sanín (2020) and Bardall et al. (2020) propose that this phenomenon, as with other forms of gendered political violence, may be rooted in the (conscious or unconscious) wish of abusers to 'punish' women who operate in a traditionally hegemonic male domain. Bardall et al. (2020) extend their theoretical framework to all politicians who do not fall within the dominant cultural, ethnic, or religious group. In the current paper, the focus is on differences in online abuse based on gender and ethnic minority status. Firstly, it is expected that Dutch female politicians receive higher levels of abuse on Twitter than male politicians. Secondly, it is hypothesized that Dutch female politicians from a minority ethnic background receive higher levels of abuse than their ethnic majority counterparts. This study distinguishes between different types of abuse on Twitter, namely, toxic tweets, severely toxic tweets, profanity, insults, identity attacks, and threats. Taking into account the framework for online abuse directed at politicians (Gorrell et al., 2020), control variables include the political orientation of politicians, their prominence, and the extent to which they engage on social media.

## Method

This study received approval from the ethics review board of the Faculty of Social and Behavioural Sciences at Utrecht University. Data and code for analysis are available via the Open Science Framework: https://osf.io/vf6xt/ In line with Twitter terms of service, we can only make tweet IDs available publicly. The full dataset is available upon request.

### *Collecting Twitter data*

All tweets mentioning at least one of the Dutch political party leaders posted between 10 January 2022 (inauguration of the House of Representatives after the 2021 general elections) to 31 December 2022 were collected using the *academictwitter* R package (Barrie & Ho, 2022) and the Twitter Academic API. In addition to party leaders (*n*=18) independent second chamber members who separated from their party at some point during 2022 or 2021 (*n*=4) were also included. Some party leaders have two Twitter accounts because they also hold a ministerial post, such as the finance minister (i.e., @SigridKaag and @Minister_Fin). In those cases (*n*=3) both accounts were included for data collection. In the API call, it was further specified that only tweets in Dutch should be collected and promoted Tweets should be excluded. In total, this resulted in a sample of 1,909,844 tweets directed at 22 politicians (see Table 3), of which 9 (40.91%) are female.

*Linguistic outcomes*
The Google Perspective API was used to measure the linguistic outcomes (i.e., dependent variables). API calls were executed using the *peRspective* R package (Votta, 2021). The tool uses machine learning models to predict 'the perceived impact a comment may have on a conversation by evaluating that comment across a range of emotional concepts'[1]. At time of writing, measures of toxicity, severe toxicity, identity attacks, insults, profanity, and threats were available in Dutch, see Table 1 for a definition of each measure provided by Google. Training data for the machine learning models consists of 'millions of comments from a variety of sources, including comments from online forums such as Wikipedia and The New York Times, across a range of languages.' Precise training data for the Dutch Perspective models are not reported. Each comment in the training data was scored on aforementioned measures by 3 to 10 annotators. Probability scores range between 0 and 1, were 1 represents a tweet in which the model predicts that all annotators would agree a tweet is toxic (or severely toxic, a threat, etc.). Model performance measured by the AUC on holdout test sets in Dutch range between 0.95 for insults to 1.0 for threats, where a score of 1 represents a model which predictions are 100% correct (see also Huang & Ling, 2005). Google recommends that a score of 0.7 or higher (i.e., 7 out of 10 annotators would agree that a tweet is toxic) can be used as a threshold in social science research for considering a tweet as toxic (or severely toxic, a threat, etc.). Before obtaining the linguistic measures via the Google Perspective API, no text preprocessing (e.g., removing mentions, URLs) was conducted because the Perspective API is developed for use on social media data. Table 2 shows examples of tweets with all six measures.

**Table 1. Definitions of linguistic measures from Google Perspective**

| Measure | Definition |
|---|---|
| 1. Toxicity | A rude, disrespectful, or unreasonable comment that is likely to make people leave a discussion. |
| 2. Severe toxicity | A very hateful, aggressive, disrespectful comment or otherwise very likely to make a user leave a discussion or give up on sharing their perspective. This attribute is much less sensitive to more mild forms of toxicity, such as comments that include positive uses of curse words. |
| 3. Identity attack | Negative or hateful comments targeting someone because of their identity, including but not limited to race or ethnicity, religion, gender, nationality or citizenship, disability, age, or sexual orientation. |
| 4. Insult | Insulting, inflammatory, or negative comment towards a person or a group of people (not identity specific) |
| 5. Profanity | Swear words, curse words, or other obscene or profane language. |
| 6. Threat | Describes an intention to inflict pain, injury, or violence against an individual or group. |

*Independent, moderator and control variables*
Gender of politicians (male or female) is entered into the regression models as independent variable. Ethnic minority status (0=ethnic majority, 1=ethnic minority) is included as a control variable in Model 1 and additionally as moderator for gender in

[1] See https://developers.perspectiveapi.com/s/about-the-api-key-concepts?language=en_US

Model 2. In both models, further control variables include prominence, political position, and engagement. Prominence is measured by the Twitter follower count of each politician on 31 December 2022. Online engagement of politicians is measured through the number of tweets sent between January 10 and 31 December 2022. Both variables are log-transformed to account for skewness in the data. Political position is based on the visualization of the Dutch political landscape along two axes developed by ‘Kieskompas’. The tool is developed by independent political researchers who make use of the official viewpoints of political parties in their manifestos (for the Dutch general elections of 2021) in order to characterize them[2]. Political position consists of two variables, namely, the economic stance of the party a politician belongs to, including the categories left-wing, center, and right-wing. The second refers to a party’s cultural stance, being either conservative or progressive. For independent second chamber members who left their party, the economic and cultural stance of the party they were previously a member of was used. See Appendix A for an overview and further justification of the political stance variables.

*Analytical approach*
Two general linear models are constructed for each of the six linguistic outcomes. Model 1 tests for a relationship between gender and the linguistic outcome, while controlling for ethnic minority status, follower count, economic stance, cultural stance, and number of tweets. In Model 2 for each outcome, the interaction between gender and ethnic minority status is entered. To account for nesting in the data (i.e., tweets are nested within politicians), cluster-robust standard errors are computed using the *miceadds* R package (Robitzsch & Grund, 2023). Confidence intervals reported are based on the clustered standard errors. Due to skewness in the data, all linguistic outcomes are log-transformed. In order to calculate the percent change in linguistic outcomes, regression coefficients *b* can be interpreted as follows: $100 * (e^b – 1)$.

Statistically significant interactions in Model 2 are further examined through pairwise comparisons of estimates for all possible combinations of gender and ethnic minority status. The *marginaleffects* R package (Arel-Bundock, 2023) is used to obtain the estimates for tweets directed at ethnic majority female, ethnic minority female, ethnic majority male, and ethnic minority male politicians. Control variables are set to their mean for continuous variables (i.e., number of followers and number of tweets) or to the reference levels used in the original models for categorical variables (i.e., cultural and economic stance). For each comparison, confidence intervals based on cluster-robust standard errors are computed.

Finally, a robustness check is performed. To verify whether regression results are not driven by the scores of an individual politician, the procedure above is repeated for each linguistic outcome, with the tweets of individual politicians who scored the highest on the respective measure removed from the data.

---

[2] See https://www.kieskompas.nl/nl/de-kieskompas-methode/ for a description of the method (in Dutch)

**Table 2. Examples of Tweets and Linguistic Outcome Measures**

| Dutch (original) | English | Gender | Toxicity | Severe toxicity | Identity attack | Insult | Pro-fanity | Threat |
|---|---|---|---|---|---|---|---|---|
| hang jezelf aub op, vieze kanker jood. Je verneukt heel nederland, hoop dat je word doorzeefd met kogels | Please hang yourself, dirty "cancer"* jew. You're ruining the Netherlands, hope you get riddled with bullets | M | 0.96 | 0.82 | 0.84 | 0.93 | 0.95 | 0.86 |
| Fuck you smerige kankerhoer, jij staat in je uppie. Helemaal alleen. Je bent een vieze landverrader. Een nsber. Een harteloos kankerwijf!!! | Fuck you dirty "cancer" whore, you're all alone. Completely alone. You are a dirty traitor to this country. An NSB'er**. A heartless "cancer" woman. | F | 0.95 | 0.82 | 0.40 | 0.93 | 0.96 | 0.28 |
| Je bent een gore kanker nazi, een loser die nooit iets zal bereiken, de letterlijke reetkanker van de samenleving. | You are a dirty "cancer" nazi, a loser who will never achieve anything, you are the "anal cancer" of society. | M | 0.95 | 0.82 | 0.59 | 0.94 | 0.94 | 0.19 |
| Smerige KutHoer,,, kogel in je kop moet je krijgen | Dirty "cunt" "cancer" whore,,, you should get a bullet in the head | F | 0.93 | 0.82 | 0.11 | 0.87 | 0.89 | 0.85 |
| ook zoiets,,,kut kanker makro,,,kom maar op ,,,9 mm ligt klaar schiet je zo overhoop,,,en daarna vermoord ik je hele fam tot in de 7e gen,.. | Again,,, moroccan,,, come on,,, 9 mm is ready to shoot you to pieces,,, and after that I kill you and your whole family to the 7th gen,.. | M | 0.93 | 0.82 | 0.32 | 0.80 | 0.90 | 0.90 |
| Rot jij maar lekker op naar je vaderland met je klote islam. | Piss off to your homecountry with your damn islam. | M | 0.92 | 0.82 | 0.85 | 0.80 | 0.76 | 0.35 |
| Domme muts ben je. Serieus wat een achtelijk schijtwijf. Geen andere woorsen voor deze domme domme domme opmerking van je. | Stupid bimbo you are. Seriously what a retarded "crap" woman. No other words for this stupid stupid stupid remark of yours. | F | 0.89 | 0.66 | 0.06 | 0.88 | 0.84 | 0.02 |
| Ik doe straks een strik om je nek heen 🖕🖕🖕 | I'm going to put a noose around your neck later 🖕🖕🖕 | M | 0.83 | 0.71 | 0.03 | 0.63 | 0.42 | 0.79 |

*Note.* English translations by the author. *Cancer (and other diseases) is a common swear word in Dutch. ** NSB was the Dutch national socialist movement in the 1930-1940s.

## Results

*Descriptive results*

Table 3 shows the descriptive statistics separated by gender. Univariate t-tests show that male politicians have significantly more followers, whereas female politicians on average have a higher level of engagement, measured by the number of tweets. Comparing mean scores for all linguistic outcomes, male politicians score significantly higher than female politicians on all measures. Nevertheless, all measures of abuse can be considered relatively low, given that the maximum score is 1. The table also shows the percentage of tweets by gender that can be classified as toxic, severely toxic etc., (i.e., the tweet received a score ≥ 0.7 for the measure). Again, a small percentage of tweets can be considered abusive, ranging between 0.01% of tweets directed at female politicians classified as a threat, to 2.45% of tweets directed at male politicians classified as an insult. Chi-square tests indicated that there were significant associations with gender for each measure, with male politicians receiving more abusive tweets across all measures.

**Table 3. Descriptive Statistics by Gender**

| | **Female** | | | **Male** | | |
|---|---|---|---|---|---|---|
| | **M** | **SD** | **%** | **M** | **SD** | **%** |
| Follower count | 129,987 | 66,840 | | 565,585* | 496,059 | |
| Number of tweets | 4,522* | 3,665 | | 2,001 | 1,514 | |
| **Linguistic outcomes** | | | | | | |
| Toxicity | 0.16 | 0.18 | 1.31 | 0.19* | 0.19 | 1.81* |
| Severe toxicity | 0.05 | 0.12 | 0.16 | 0.07* | 0.14 | 0.24* |
| Identity attack | 0.04 | 0.09 | 0.05 | 0.05* | 0.11 | 0.11* |
| Insult | 0.15 | 0.20 | 1.66 | 0.19* | 0.22 | 2.45* |
| Profanity | 0.10 | 0.15 | 1.47 | 0.12* | 0.16 | 1.59* |
| Threat | 0.02 | 0.05 | 0.01 | 0.02* | 0.06 | 0.02* |

*Note.* *$p$<0.001

In Table 4, the linguistic outcomes are separated by individual politicians, showing the number of tweets out of 100 directed at each politician that is classified as toxic, severely toxic, an identity attack, an insult, profanity, or a threat (i.e., ≥ 0.7). Mark Rutte, the male prime minister, receives the highest proportion of toxic tweets, with 3.58 out of 100 tweets directed at him being classified as such. Sigrid Kaag, the female vice prime minister and finance minister, receives the second highest proportion of toxic tweets (3.41/100), but the highest proportion of *severely* toxic tweets (0.53/100). Identity attacks are less common, but here Farid Azarkan (0.62/100), Nilüfer Gundogan (0.22/100), and Sylvana Simons (0.14/100), all with an ethnic minority background, receive the most. Profanity is most commonly found in tweets directed at Liane Den Haan (4.30/100)[3], Sigrid Kaag (3.28/100) and Mark Rutte (2.90/100) Finally, although threats are the least common, Geert Wilders[4] receives the most (0.04/100), followed by Mark Rutte (0.03/100) and Sigrid Kaag (0.02/100).

[3] This high score is somewhat misleading, since Liane den Haan on 13/7/2022 posted a tweet condemning the use of #kutland ('shit country'), and many tweets in reply to this included the expletive 'kut', hence the high score on profanity for this politician.

[4] Geert Wilders is the leader of a political party that is regarded as xenophobic and anti-Islam (Witteveen, 2017). Manual inspection of tweets directed at Geert Wilders show that they frequently contain threats directed at others (mainly Muslims) rather than at the politician himself.

*Regression models*
Results for the regression analysis are shown in Table 5. Contrary to expectations, tweets directed at female politicians score lower (as indicated by the negative regression coefficient) on (1) *toxicity,* (2) *severe toxicity,* (3) *identity attacks*, (4) *insults,* and (5) *profanity* compared to their male counterparts when controlling for number of followers, ethnic minority status, economic and cultural stance, and number of tweets. Recall that the dependent variables were log-transformed. This means, for example, that female politicians score $100*(e^{-0.29} – 1) = 25.2\%$ *lower* on toxicity than male politicians, and $100*(e^{-0.31} – 1) = 26.7\%$ *lower* on severe toxicity, and so on. No significant differences between male and female politicians were found for the levels of (6) *threats*. Significant effects for ethnic minority status can be observed in Model 1 for the measures of profanity and threats, with ethnic minority politicians receiving 18.5% and 8.3% higher scores on these measures, respectively.

When introducing the interaction between gender and ethnic minority status in Model 2 for each measure, a significant positive interaction for (2) *severe toxicity,* (3) *identity attacks,* (5) *profanity,* and (6) *threats* can be observed. No significant interaction effect between gender and ethnic minority status was found for (1) *toxicity* nor for (4) *insults.* The models explain only a small amount of variance for each linguistic outcome, ranging from 2.6% of variance for threats to 4.7% for severe toxicity (see pseudo $R^2$ for Model 2 in both cases).

Interaction plots for (2) *severe toxicity,* (3) *identity attacks,* (5) *profanity,* and (6) *threats* are shown in Figure 1, with pairwise comparisons between each group reported in Appendix B. Examining the plot for (2) *severe toxicity,* tweets directed at ethnic minority female politicians score higher than tweets directed at majority female politicians ($p<0.001$, see Appendix B) but not significantly higher than tweets directed at both ethnic minority and majority male groups. For the measures of (3) *identity attacks* and (5) *profanity*, a similar pattern emerges. That is, a significant difference between tweets directed at ethnic majority and minority female politicians is found, but tweets for ethnic minority female politicians do not score any different from tweets mentioning male politicians. For the measure of (6) *threats*, tweets directed at minority ethnic female politicians score the highest. The estimate for this group is significantly higher than that for tweets directed at female ethnic majority, male ethnic majority, and male ethnic minority politicians ($p<0.001$).

The results for the robustness check are shown in Appendix C. For each linguistic outcome, the tweets directed at the politician that scored the highest on that measure (as shown in Table 4) were removed from the data and both models were constructed. Results show similar effects in terms of directionality and statistical significance as those shown in Table 5.

**Table 4. Linguistic Outcomes per Politician**

| Politician | Gender | Minority ethnic | Toxic | Severely toxic | Identity attack | Insult | Profanity | Threat |
|---|---|---|---|---|---|---|---|---|
| Mark Rutte | M | N | **3.58** | 0.49 | 0.06 | **4.64** | 2.90 | 0.03 |
| Sigrid Kaag | F | N | 3.41 | **0.53** | 0.12 | 3.91 | 3.28 | 0.02 |
| Jesse Klaver | M | Y | 3.21 | 0.38 | 0.17 | 4.32 | 2.38 | 0.01 |
| Nilüfer Gundogan | F | Y | 2.86 | 0.49 | 0.22 | 3.75 | 2.05 | 0.01 |
| Wopke Hoekstra | M | N | 2.25 | 0.22 | 0.05 | 3.15 | 1.84 | 0.01 |
| Sylvana Simons | F | Y | 2.17 | 0.33 | 0.14 | 2.63 | 1.88 | 0.00 |
| Laurens Dassen | M | N | 2.14 | 0.21 | 0.06 | 2.56 | 2.05 | 0.01 |
| Attje Kuiken | F | N | 1.95 | 0.11 | 0.01 | 2.84 | 1.75 | 0.00 |
| Liane den Haan | F | N | 1.90 | 0.12 | 0.01 | 2.00 | **4.30** | 0.00 |
| Farid Azarkan | M | Y | 1.61 | 0.50 | **0.62** | 1.45 | 1.40 | 0.01 |
| Thierry Baudet | M | N | 1.58 | 0.12 | 0.02 | 2.56 | 1.57 | 0.01 |
| Geert Wilders | M | N | 1.57 | 0.31 | 0.27 | 1.79 | 1.43 | **0.04** |
| Joost Eerdmans | M | N | 1.27 | 0.02 | 0.02 | 2.04 | 1.31 | 0.00 |
| Lilianne Ploumen | F | N | 1.22 | 0.06 | 0.02 | 1.56 | 1.18 | 0.00 |
| Esther Ouwehand | F | N | 1.22 | 0.11 | 0.03 | 1.62 | 1.54 | 0.01 |
| Wybren van Haga | M | N | 0.86 | 0.06 | 0.02 | 1.54 | 0.99 | 0.00 |
| Gertjan Segers | M | N | 0.79 | 0.09 | 0.07 | 1.11 | 0.67 | 0.00 |
| Caroline van der Plas | F | N | 0.32 | 0.03 | 0.01 | 0.49 | 0.64 | 0.00 |
| Lilian Marijnissen | F | N | 0.31 | 0.02 | 0.02 | 0.46 | 0.68 | 0.00 |
| Kees van der Staaij | M | N | 0.27 | 0.06 | 0.02 | 0.56 | 0.56 | 0.00 |
| Pieter Omtzigt | M | N | 0.27 | 0.02 | 0.01 | 0.43 | 0.50 | 0.00 |
| Martin van Rooijen | M | N | 0.24 | 0.02 | 0.01 | 0.36 | 0.32 | 0.00 |

*Note.* Number of tweets out of 100 tweets directed at a politician that can be classified as each form of abuse are shown. Highest score per measure in bold.

## Table 5. Regression results for linguistic outcomes per model

| | Linguistic Outcome | | | | | | | | | | | |
|---|---|---|---|---|---|---|---|---|---|---|---|---|
| | 1. Toxicity | | 2. Severe toxicity | | 3. Identity attack | | 4. Insult | | 5. Profanity | | 6. Threat | |
| **Model** | (1) | (2) | (1) | (2) | (1) | (2) | (1) | (2) | (1) | (2) | (1) | (2) |
| **Gender** | -0.29** | -0.31*** | -0.31*** | -0.35*** | -0.38*** | -0.42*** | -0.26* | -0.27** | -0.19** | -0.21*** | -0.03 | -0.04** |
| (ref=M) | [-0.47, -0.12] | [-0.47, -0.15] | [-0.48, -0.14] | [-0.49, -0.21] | [-0.51, -0.25] | [-0.51, -0.33] | [-0.45, -0.06] | [-0.46, -0.09] | [-0.32, -0.07] | [-0.32, -0.10] | [-0.07, 0.01] | [-0.08, -0.01] |
| **Ethnic minority** | 0.10 | 0.03 | 0.23 | 0.06 | 0.08 | -0.11 | 0.13 | 0.06 | 0.17* | 0.08 | 0.08* | 0.02 |
| (ref=maj.) | [-0.12, 0.32] | [-0.25, 0.32] | [-0.01, 0.47] | [-0.14, 0.26] | [-0.18, 0.34] | [-0.26, 0.04] | [-0.10, 0.37] | [-0.25, 0.37] | [0.01, 0.33] | [-0.10, 0.26] | [0.00, 0.16] | [-0.03, 0.06] |
| **Economic** | -0.06 | -0.10 | -0.17 | -0.26 | -0.20* | -0.30** | -0.07 | -0.11 | -0.07 | -0.11 | -0.07 | -0.10** |
| (ref=left) *Center* | [-0.33, 0.21] | [-0.39, 0.19] | [-0.41, 0.08] | [-0.51, 0.00] | [-0.40, 0.00] | [-0.52, -0.07] | [-0.36, 0.21] | [-0.42, 0.20] | [-0.25, 0.12] | [-0.31, 0.08] | [-0.14, 0.01] | [-0.17, -0.03] |
| *Right-wing* | 0.39* | 0.36* | 0.41* | 0.33* | 0.33* | 0.25 | 0.36* | 0.32 | 0.36** | 0.32** | 0.12* | 0.09* |
| | [0.07, 0.71] | [0.03, 0.69] | [0.09, 0.73] | [0.01, 0.66] | [0.05, 0.61] | [-0.05, 0.54] | [0.01, 0.70] | [-0.03, 0.68] | [0.12, 0.59] | [0.08, 0.55] | [0.03, 0.21] | [0.00, 0.18] |
| **Cultural** | 0.20 | 0.19 | 0.12 | 0.10 | 0.23 | 0.21 | 0.16 | 0.15 | 0.08 | 0.07 | -0.04 | -0.05 |
| (ref=cons.) | [-0.14, 0.53] | [-0.15, 0.52] | [-0.25, 0.48] | [-0.26, 0.46] | [-0.09, 0.55] | [-0.10, 0.52] | [-0.20, 0.51] | [-0.20, 0.50] | [-0.18, 0.34] | [-0.19, 0.33] | [-0.14, 0.06] | [-0.14, 0.05] |
| **Tweets** | -0.21*** | -0.21*** | -0.30*** | -0.31*** | -0.27*** | -0.28*** | -0.22*** | -0.22*** | -0.19*** | -0.20*** | -0.12*** | -0.12*** |
| (log) | [-0.30, -0.12] | [-0.31, -0.12] | [-0.40, -0.19] | [-0.40, -0.21] | [-0.36, -0.18] | [-0.36, -0.20] | [-0.31, -0.12] | [-0.32, -0.12] | [-0.26, -0.12] | [-0.26, -0.13] | [-0.15, -0.09] | [-0.15, -0.09] |
| **Followers** | -0.06 | -0.05 | -0.05 | -0.03 | -0.02 | -0.01 | -0.05 | -0.04 | -0.06* | -0.05 | 0.01 | 0.02 |
| (log) | [-0.14, 0.03] | [-0.14, 0.04] | [-0.13, 0.04] | [-0.12, 0.05] | [-0.10, 0.05] | [-0.08, 0.07] | [-0.13, 0.04] | [-0.13, 0.05] | [-0.12, 0.00] | [-0.11, 0.01] | [-0.02, 0.04] | [-0.01, 0.04] |
| **Gender* ethnicity** | | 0.19 | | 0.46*** | | 0.50** | | 0.19 | | 0.24** | | 0.18*** |
| | | [-0.11, 0.50] | | [0.22, 0.70] | | [0.20, 0.80] | | [-0.14, 0.52] | | [0.07, 0.42] | | [0.13, 0.22] |
| **Pseudo $R^2$** | 0.032 | 0.033 | 0.045 | 0.047 | 0.038 | 0.040 | 0.031 | 0.031 | 0.031 | 0.032 | 0.025 | 0.026 |

*Note.* Unstandardized regression coefficients *b* and confidence intervals based on cluster-robust standard errors. Linguistic outcomes were log-transformed. * $p < .05$; ** $p < .01$; *** $p < .001$.

**Figure 1. Interaction Plots**

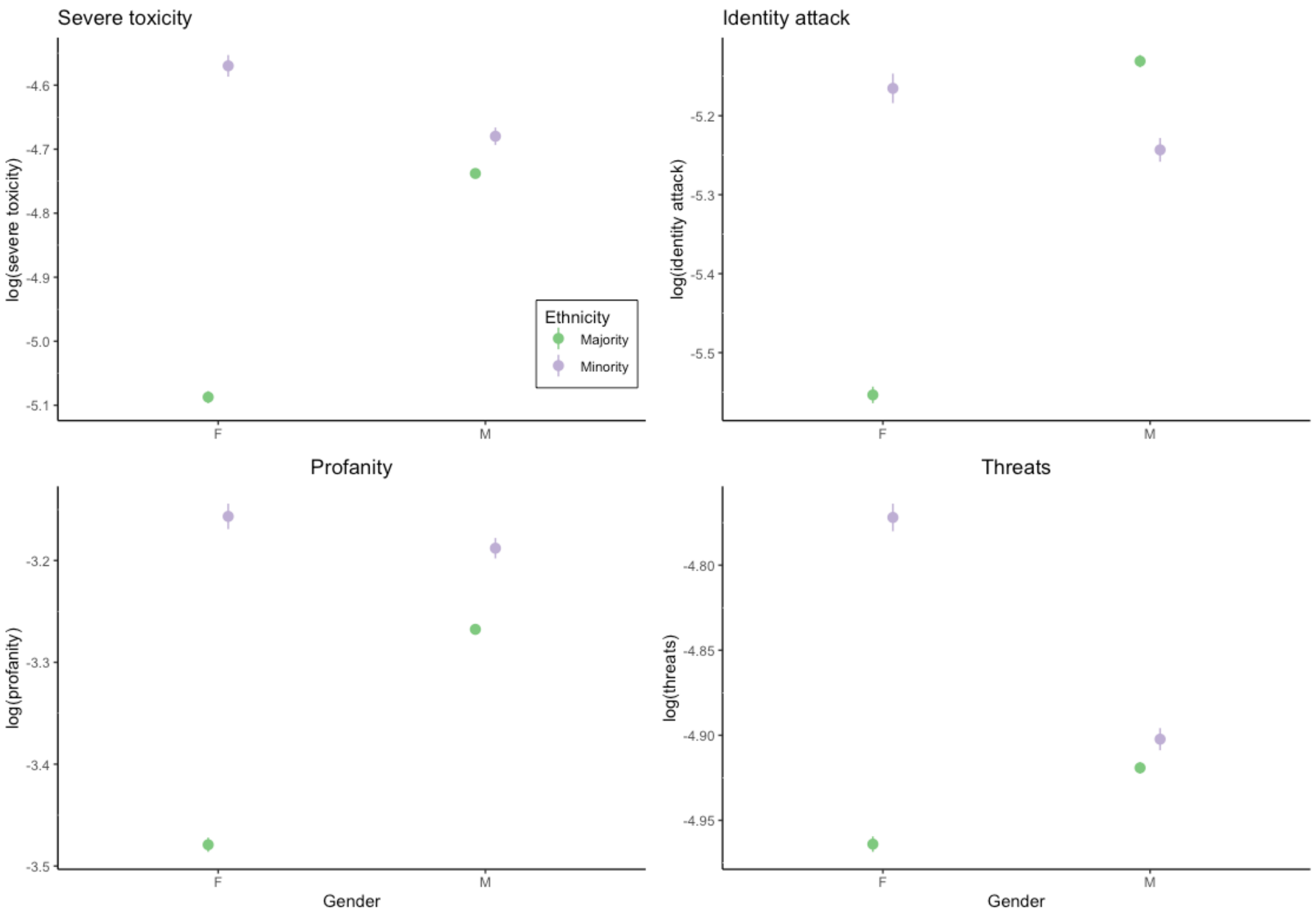

*Note.* Estimates are based on a model where number of followers and number of tweets are set to their mean, and where the original reference levels are used for economic (i.e., left-wing) and cultural (i.e., conservative) stance. Lines around point estimates represent error bars.

**Discussion**

The results of this study paint a mixed picture of the abuse Dutch politicians receive on Twitter. While the overall levels of abuse can be considered low, descriptive results did show the disproportionate nature of abuse. Stark differences between individual politicians were found, with prominent politicians receiving the highest levels of (severe) toxicity, while identity attacks were primarily directed at ethnic minority politicians.

The key findings of this paper are as follows. Female politicians receive lower levels of most forms of abuse compared to their male counterparts, including toxicity, severe toxicity, identity attacks, insults, and profanity. No statistically significant difference between male and female politicians was found for the measure of threats. Ethnic minority status significantly increases the level of severe toxicity, identity attacks, profanity, and threats female politicians receive. For the measures of severe toxicity, identity attacks, and profanity, scores for ethnic minority female politicians are heightened to the level of that of tweets directed at both ethnic minority and majority male politicians. Importantly, for the measure of threats only, tweets directed at ethnic minority female politicians score the highest when compared to tweets directed at all other groups. In short, no support for the hypothesis that female politicians receive higher levels of abuse than male politicians on Twitter is found. The results offer partial support for the hypothesis stating that ethnic minority status increases the level of abuse female politicians receive. This holds for the measures of severe toxicity, identity attacks, profanity, and threats only.

Perhaps the most striking result of this study is that being an ethnic minority woman in Dutch politics appears to be associated with the highest levels of threats, arguably the most severe form of abuse. While ethnic minority female politicians are not impacted more strongly than male politicians on the other forms of abuse measured in this study, it is clear that they are generally worse off than their ethnic majority female colleagues. Looking back at the theoretical frameworks for gendered political violence (Bardall et al., 2020; Krook & Sanín, 2020), a few remarks can be made. Expectations were derived from the idea that online abuse of female politicians can be explained by the motivation to preserve politics as hegemonic male domain. Some support for this explanation was found, albeit not in the way that was expected. Tweets directed at women politicians did not score significantly higher on all of measures of abuse. However, the expected pattern of results was obtained for female politicians with an ethnic minority background for some of the outcome measures. Therefore, the possibility still exists that abuse (and threats in particular) are used to ‘punish’ ethnic minority women for their participation in the political arena, in line with preserving politics as a hegemonic male domain. Although significant differences were found based on gender in combination with ethnic minority status, it cannot be concluded that the abuse expressed in this sample of tweets is *explicitly* motivated by these characteristics. Based on the probability scores for abuse, we cannot establish whether gender and ethnicity attributes of politicians are, for example, mentioned in a derogatory way. The measure of ‘identity attacks’ does concern abuse based on someone’s identity, but the probability score does not show whether the attack is based on gender, ethnicity, religion, sexual orientation, or some other characteristic. In future research, more in-depth analysis of the content of tweets will be necessary.

In short, results appear to be more in line with previous work suggesting that male politicians receive higher levels of general abuse online (Gorrell et al., 2020; Ward & McLoughlin, 2020). At the same time, some of the results align with earlier investigations showing that female ethnic minority politicians are disproportionately affected by online abuse (Amnesty International UK, 2017; Esposito & Breeze, 2022; Saris & van de Ven, 2021). However, direct comparisons of current results with previous studies are difficult since linguistic measures and analytical approaches were different. For instance, average levels of abuse directed at female politicians in this study vary between 0.01% for threatening and 1.66% for insulting tweets, which can be considered sufficiently lower than the 10% of tweets classified as abusive or threatening by Saris & van de Ven (2021) in their Dutch sample. However, due to differences in definitions of abuse and data collection procedures, these findings cannot be directly compared, and thus it would not be appropriate to say that levels of abuse directed at female politicians have decreased.

The theoretical frameworks leveraged in this study to explain abuse directed at female politicians cannot be utilized to understand the high(er) levels of abuse directed at male politicians in the data. Alternative explanations can perhaps be found in the Gorell et al. (2020) four factor framework, which considers prominence, event surge, engagement, and identity as main drivers of online abuse of politicians. The possibility exists that male politicians receive more abuse than their female counterparts due to their prominence or engagement with the public, and that these factors were not adequately controlled for with the measures of Twitter follower count and number of tweets, respectively. Furthermore, the current study did not take into account so-called ‘event surge’, that is, specific political or media events that may have led to high volumes of abuse. Another possible explanation may lie in the linguistic measures utilized in this study. For instance, the measure of ‘identity attacks’ is defined by Google Perspective as any attack based on someone’s gender, ethnicity, race, or other personal characteristics. As a consequence, one might expect ethnic minority politicians and women in particular to receive a high level of such messages. Surprisingly, majority ethnic male politicians received the highest level of identity attacks in this study. One explanation for this could be that leaders of known xenophobic parties[4] are included in the majority male group. In some cases, it was apparent that while a tweet was directed at a politician, it included (racist) abuse or threats directed at another individual or group and not at the politician him/herself. As mentioned above, the tools used for linguistic measurement in this study do not provide specific information on the target of abuse or the personal characteristics of a target motivating an identity attack. In addition to future qualitative examination of tweets to determine explicit motivations and targets, the application of more sophisticated target detection algorithms could offer some further insight on this matter.

Some additional imitations to this study need to be considered. First, the data collection procedure only retrieved tweets that directly mentioned a politician by their Twitter username. Therefore, tweets in which a politician was simply named or implicitly referred to were not included in the analysis. The possibility exists that such tweets are more or less abusive than what was captured in this dataset. Second, on October 27, 2022, Elon Musk became Twitter’s new owner and CEO. Various reports have stated that content moderation has suffered or at least changed as a result of this (Paul & Dang, 2022), as Musk worked to instate his “free speech” agenda on the platform (Zakrzewski et al., 2022). These events may have had an effect on the results observed in this paper. At the same time, the possibility also exists that the levels of abuse posted on Twitter were underestimated due to content moderation, with us failing to capture tweets that were removed from the platform as a result of this. All in all, these factors urge us to interpret (the magnitude and direction of) the results presented with some caution. Future research may also analyze the data from a timeseries perspective, in order to assess the potential effect of the Musk acquisition or specific political events. Additionally, free access to the Twitter Academic API has been curtailed since early 2023 (Calma, 2023), complicating future replication efforts or additional data collection. Finally, a third-party tool such as the Google Perspective API does not allow for fully transparent linguistic measurements. That is, the documentation does not specify the precise training data used for the Dutch models and possible bias as a result of this. Furthermore, since the tool merely produces a probability score for each measure of abuse, it was not possible to scrutinize the precise features the models use to predict the levels of toxicity, identity attacks, and other measures.

## Conclusion

Politicians around the world, including in the Netherlands, have reported a significant and rising trend of online abuse. This paper set out to add to growing field of empirical inquiry into gender differences in abuse received by politicians online, while taking into account the additional impact of ethnic minority status. Six different forms of abuse were measured in a full year of Twitter data. Contrary to the expectations set out at the beginning of this study, male politicians receive higher levels of most forms of abuse. At the same time, female ethnic minority politicians are more severely affected than female ethnic majority politicians on some forms of abuse. Notably, female ethnic minority politicians receive the highest level of threats, the most severe form of abuse measured in this study. While it cannot be said how many of such threats may lead to actual physical violence, the mere prevalence and nature of abuse in the online domain are reportedly already enough to have women leave or refrain from participating in the political arena. Therefore, it is important that we continue to invest in understanding and combatting this phenomenon.

**Appendix A: Political stance measures**

In order to determine political stance of politicians, a visualization of the Dutch political landscape for the 2021 general elections developed by the independent organization ‘Kieskompas’ was used (see https://tweedekamer2021.kieskompas.nl/nl/). In the visualization, each political party is placed along the horizontal axis of socio-economic viewpoints (left to right-wing) and the vertical axis of cultural (progressive to conservative) viewpoints. All political parties left of the mid-point of the horizontal axis were coded as left-wing in terms of socio-economic viewpoints, all parties to the right of the mid-point as right-wing, and those who intersected with the mid-point of the horizontal axis were coded as ‘center’. We similarly considered all political parties above the mid-point of the vertical axis as progressive, and those below as conservative. No political parties crossed the mid-point of the vertical axis for cultural viewpoints. The below table shows the coding for each politician.

| Party leader | Political Party | Economic | Cultural |
|---|---|---|---|
| Mark Rutte | VVD | Right-wing | Conservative |
| Sigrid Kaag | D66 | Center | Progressive |
| Jesse Klaver | GroenLinks | Left-wing | Progressive |
| Nilüfer Gundogan | Independent (former VOLT) | Center | Progressive |
| Wopke Hoekstra | CDA | Center | Conservative |
| Sylvana Simons | BIJ1 | Left-wing | Progressive |
| Laurens Dassen | VOLT | Center | Progressive |
| Liane den Haan | Independent (former 50Plus) | Left-wing | Progressive |
| Farid Azarkan | DENK | Left-wing | Progressive |
| Thierry Baudet | Forum voor Democratie | Right-wing | Conservative |
| Geert Wilders | PVV | Left-wing | Conservative |
| Joost Eerdmans | JA21 | Right-wing | Conservative |
| Lilianne Ploumen | PvdA (until 21 April 2022) | Left-wing | Progressive |
| Attje Kuiken | PvdA (from 22 April 2022) | Left-wing | Progressive |
| Esther Ouwehand | Partij voor de Dieren | Left-wing | Progressive |
| Wybren van Haga | Independent (former FvD) | Right-wing | Conservative |
| Gertjan Segers | ChristenUnie | Left-wing | Progressive |
| Caroline van der Plas | BBB | Right-wing | Conservative |
| Lilian Marijnissen | SP | Left-wing | Progressive |
| Kees van der Staaij | SGP | Right-wing | Conservative |
| Pieter Omtzigt | Independent (former CDA) | Center | Conservative |
| Martin van Rooijen | 50Plus | Left-wing | Conservative |

## Appendix B: Pairwise comparisons (contrasts)

A positive estimated difference suggests that the first group in the comparison scores higher than the second group, and vice versa. Confidence intervals are computed based on cluster-robust standard errors, with: * $p < .05$; ** $p < .01$; *** $p < .001$.

### Severe toxicity

| Contrast | Est. diff | $CI_{low}$ | $CI_{high}$ |
|---|---|---|---|
| Majority male – majority female | 0.349*** | 0.212 | 0.487 |
| Majority male – minority male | -0.058 | -0.256 | 0.14 |
| Majority male – minority female | -0.168 | -0.43 | 0.094 |
| Majority female – minority male | -0.407** | -0.655 | -0.16 |
| Majority female – minority female | -0.518*** | -0.745 | -0.29 |
| Minority male – minority female | -0.110 | -0.387 | 0.167 |

### Identity attack

| Contrast | Est. diff | $CI_{low}$ | $CI_{high}$ |
|---|---|---|---|
| Majority male – majority female | 0.423*** | 0.331 | 0.514 |
| Majority male – minority male | 0.112 | -0.037 | 0.261 |
| Majority male – minority female | 0.034 | -0.296 | 0.365 |
| Majority female – minority male | -0.31** | -0.506 | -0.115 |
| Majority female – minority female | -0.388* | -0.715 | -0.061 |
| Minority male – minority female | -0.078 | -0.395 | 0.239 |

### Profanity

| Contrast | Est. diff | $CI_{low}$ | $CI_{high}$ |
|---|---|---|---|
| Majority male – majority female | 0.211*** | 0.101 | 0.322 |
| Majority male – minority male | -0.080 | -0.256 | 0.097 |
| Majority male – minority female | -0.111 | -0.291 | 0.069 |
| Majority female – minority male | -0.291** | -0.505 | -0.077 |
| Majority female – minority female | -0.322*** | -0.475 | -0.170 |
| Minority male – minority female | -0.031 | -0.234 | 0.172 |

### Threat

| Contrast | Est. diff | $CI_{low}$ | $CI_{high}$ |
|---|---|---|---|
| Majority male – majority female | 0.045** | 0.014 | 0.076 |
| Majority male – minority male | -0.017 | -0.062 | 0.028 |
| Majority male – minority female | -0.147*** | -0.205 | -0.09 |
| Majority female – minority male | -0.062 | -0.121 | -0.002 |
| Majority female – minority female | -0.192*** | -0.241 | -0.143 |
| Minority male – minority female | -0.130*** | -0.191 | -0.070 |

## Appendix C: Robustness check

| | **Linguistic Outcome** | | | | | | | | | | | |
|---|---|---|---|---|---|---|---|---|---|---|---|---|
| | 1. Toxicity | | 2. Severe toxicity | | 3. Identity attack | | 4. Insult | | 5. Profanity | | 6. Threat | |
| **Model** | (1) | (2) | (1) | (2) | (1) | (2) | (1) | (2) | (1) | (2) | (1) | (2) |
| **Gender** | -0.32*** | -0.33*** | -0.32** | -0.38*** | -0.40*** | -0.43*** | -0.29** | -0.29** | -0.21*** | -0.24*** | -0.03 | -0.05** |
| (ref=M) | [-0.51, -0.14] | [-0.51, -0.15] | [-0.51, -0.13] | [-0.54, -0.22] | [-0.51, -0.28] | [-0.52, -0.33] | [-0.50, -0.08] | [-0.50, -0.09] | [-0.34, -0.09] | [-0.35, -0.13] | [-0.07, 0.01] | [-0.08, -0.02] |
| **Ethnic minority** | 0.11 | 0.09 | 0.24 | 0.06 | 0.11 | -0.09 | 0.14 | 0.12 | 0.15 | 0.06 | 0.08* | 0.01 |
| (ref=maj.) | [-0.12, 0.34] | [-0.22, 0.40] | [-0.01, 0.48] | [-0.14, 0.26] | [-0.18, 0.40] | [-0.25, 0.07] | [-0.11, 0.39] | [-0.23, 0.46] | [-0.02, 0.32] | [-0.12, 0.24] | [0.00, 0.16] | [-0.04, 0.05] |
| **Economic** (ref=left) *Center* | 0.05 | 0.04 | -0.19 | -0.33* | -0.20 | -0.29* | 0.04 | 0.03 | 0.00 | -0.05 | -0.07 | -0.11** |
| | [-0.24, 0.34] | [-0.29, 0.36] | [-0.41, 0.02] | [-0.59, -0.06] | [-0.40, 0.01] | [-0.52, -0.06] | [-0.26, 0.35] | [-0.32, 0.38] | [-0.20, 0.20] | [-0.25, 0.16] | [-0.15, 0.01] | [-0.18, -0.04] |
| *Right-wing* | 0.45* | 0.44* | 0.39** | 0.27 | 0.35* | 0.26 | 0.42* | 0.41* | 0.42*** | 0.38** | 0.12* | 0.09+ |
| | [0.10, 0.80] | [0.07, 0.81] | [0.11, 0.66] | [-0.03, 0.57] | [0.07, 0.63] | [-0.04, 0.56] | [0.04, 0.80] | [0.01, 0.81] | [0.17, 0.67] | [0.13, 0.63] | [0.03, 0.21] | [0.00, 0.17] |
| **Cultural** | 0.32 | 0.31 | 0.10 | 0.06 | 0.23 | 0.21 | 0.28 | 0.27 | 0.20 | 0.19 | -0.04 | -0.05 |
| (ref=cons.) | [-0.06, 0.70] | [-0.08, 0.70] | [-0.23, 0.43] | [-0.29, 0.40] | [-0.08, 0.54] | [-0.09, 0.51] | [-0.12, 0.67] | [-0.14, 0.68] | [-0.11, 0.50] | [-0.11, 0.49] | [-0.14, 0.06] | [-0.14, 0.05] |
| **Tweets** | -0.16*** | -0.17** | -0.29*** | -0.29*** | -0.27*** | -0.28*** | -0.17** | -0.17** | -0.17*** | -0.17*** | -0.12*** | -0.12*** |
| (log) | [-0.26, -0.07] | [-0.27, -0.06] | [-0.42, -0.16] | [-0.41, -0.17] | [-0.36, -0.18] | [-0.36, -0.20] | [-0.27, -0.07] | [-0.28, -0.06] | [-0.25, -0.09] | [-0.25, -0.10] | [-0.15, -0.09] | [-0.15, -0.10] |
| **Followers** | -0.08* | -0.08 | -0.05 | -0.03 | -0.03 | -0.01 | -0.07 | -0.07 | -0.05 | -0.04 | 0.01 | 0.02 |
| (log) | [-0.16, 0.00] | [-0.17, 0.01] | [-0.13, 0.04] | [-0.11, 0.06] | [-0.11, 0.04] | [-0.09, 0.07] | [-0.16, 0.01] | [-0.17, 0.02] | [-0.11, 0.01] | [-0.10, 0.02] | [-0.02, 0.04] | [-0.01, 0.05] |
| **Gender* ethnicity** | | 0.06 | | 0.50*** | | 0.48** | | 0.05 | | 0.25** | | 0.19*** |
| | | [-0.30, 0.41] | | [0.28, 0.71] | | [0.17, 0.78] | | [-0.34, 0.45] | | [0.07, 0.43] | | [0.14, 0.23] |
| **Pseudo $R^2$** | 0.027 | 0.027 | 0.045 | 0.046 | 0.039 | 0.041 | 0.026 | 0.026 | 0.032 | 0.033 | 0.028 | 0.029 |
| **N** | 1,720,403 | | 1,807,688 | | 1,877,889 | | 1,720,403 | | 1,890,893 | | 1,674,094 | |
| **Politician removed** | *Mark Rutte* | | *Sigrid Kaag* | | *Farid Azarkan* | | *Mark Rutte* | | *Liane den Haan* | | *Geert Wilders* | |

*Note.* For each linguistic outcome, the politician with the highest score on that outcome (see Table 4) is removed from the data. Unstandardized regression coefficients *b* and cluster-robust standard errors are reported. Linguistic outcomes were log-transformed. * $p < .05$; ** $p < .01$; *** $p < .001$.